\documentclass[acmtog]{acmart}
\renewcommand\footnotetextcopyrightpermission[1]{} % removes footnote with conference information in first column
\AtBeginDocument{%
  \providecommand\BibTeX{{%
    \normalfont B\kern-0.5em{\scshape i\kern-0.25em b}\kern-0.8em\TeX}}}

\setcopyright{acmcopyright}
\copyrightyear{2019}
\usepackage{amsmath} % assumes amsmath package installed
\usepackage{amssymb}  % assumes amsmath package installed
\usepackage{algorithm2e}
\usepackage{graphicx}
\usepackage{subcaption}
\usepackage{caption}
\usepackage{multirow}
\usepackage{multicol}
\begin{document}

%%
%% The "title" command has an optional parameter,
%% allowing the author to define a "short title" to be used in page headers.
\title{Deep Reinforcement Learning for Single-Shot Diagnosis and Adaptation in Damaged Robots}

%%
%% The "author" command and its associated commands are used to define
%% the authors and their affiliations.
%% Of note is the shared affiliation of the first two authors, and the
%% "authornote" and "authornotemark" commands
%% used to denote shared contribution to the research.
% \author{Ben Trovato}
% \authornote{Both authors contributed equally to this research.}
% \email{trovato@corporation.com}
% \orcid{1234-5678-9012}
% \author{G.K.M. Tobin}
% \authornotemark[1]
% \email{webmaster@marysville-ohio.com}
% \affiliation{%
%   \institution{Institute for Clarity in Documentation}
%   \streetaddress{P.O. Box 1212}
%   \city{Dublin}
%   \state{Ohio}
%   \postcode{43017-6221}
% }

\author{Shresth Verma}
\affiliation{%
  \institution{ABV-Indian Institute of Information Technology and Management, Gwalior}}
\email{vermashresth@gmail.com}
\orcid{0000-0003-0370-5471}

\author{Haritha S. Nair}
\affiliation{%
  \institution{ABV-Indian Institute of Information Technology and Management, Gwalior}}
  \email{haritha1313@gmail.com
}

% \author{Aparna Patel}
% \affiliation{%
%  \institution{Rajiv Gandhi University}
%  \streetaddress{Rono-Hills}
%  \city{Doimukh}
%  \state{Arunachal Pradesh}
%  \country{India}}

% \author{Huifen Chan}
% \affiliation{%
%   \institution{Tsinghua University}
%   \streetaddress{30 Shuangqing Rd}
%   \city{Haidian Qu}
%   \state{Beijing Shi}
%   \country{China}}

% \author{Charles Palmer}
% \affiliation{%
%   \institution{Palmer Research Laboratories}
%   \streetaddress{8600 Datapoint Drive}
%   \city{San Antonio}
%   \state{Texas}
%   \postcode{78229}}
% \email{cpalmer@prl.com}

\author{Gaurav Agarwal}
\affiliation{%
  \institution{ABV-Indian Institute of Information Technology and Management, Gwalior}}
\email{gaurava05@gmail.com}

\author{Joydip Dhar}
\affiliation{\institution{ABV-Indian Institute of Information Technology and Management, Gwalior}}
\email{jdhar@iiitm.ac.in}

\author{Anupam Shukla}
\affiliation{\institution{ABV-Indian Institute of Information Technology and Management, Gwalior}}
\email{anupamshukla@iiitm.ac.in}

%%
%% By default, the full list of authors will be used in the page
%% headers. Often, this list is too long, and will overlap
%% other information printed in the page headers. This command allows
%% the author to define a more concise list
%% of authors' names for this purpose.
% \renewcommand{\shortauthors}{Shresth, et al.}

%%
%% The abstract is a short summary of the work to be presented in the
%% article.
\begin{abstract}
Robotics has proved to be an indispensable tool in many industrial as well as social applications, such as warehouse automation, manufacturing, disaster robotics, etc. In most of these scenarios, damage to the agent while accomplishing mission-critical tasks can result in failure. To enable robotic adaptation in such situations, the agent needs to adopt policies which are robust to a diverse set of damages and must do so with minimum computational complexity. We thus propose a damage aware control architecture which diagnoses the damage prior to gait selection while also incorporating domain randomization in the damage space for learning a robust policy. To implement damage awareness, we have used a Long Short Term Memory based supervised learning network which diagnoses the damage and predicts the type of damage. The main novelty of this approach is that only a single policy is trained to adapt against a wide variety of damages and the diagnosis is done in a single trial at the time of damage.
\end{abstract}

%%
%% The code below is generated by the tool at http://dl.acm.org/ccs.cfm.
%% Please copy and paste the code instead of the example below.
%%
\begin{CCSXML}
<ccs2012>
<concept>
<concept_id>10003752.10010070.10010071.10010261</concept_id>
<concept_desc>Theory of computation~Reinforcement learning</concept_desc>
<concept_significance>500</concept_significance>
</concept>
<concept>
<concept_id>10010520.10010553.10010554.10010556</concept_id>
<concept_desc>Computer systems organization~Robotic control</concept_desc>
<concept_significance>500</concept_significance>
</concept>
<concept>
<concept_id>10010147.10010257.10010258.10010259.10010263</concept_id>
<concept_desc>Computing methodologies~Supervised learning by classification</concept_desc>
<concept_significance>300</concept_significance>
</concept>
<concept>
<concept_id>10010147.10010257.10010258.10010261.10010272</concept_id>
<concept_desc>Computing methodologies~Sequential decision making</concept_desc>
<concept_significance>300</concept_significance>
</concept>
<concept>
<concept_id>10010147.10010257.10010258.10010262</concept_id>
<concept_desc>Computing methodologies~Multi-task learning</concept_desc>
<concept_significance>300</concept_significance>
</concept>
</ccs2012>
\end{CCSXML}

\ccsdesc[500]{Theory of computation~Reinforcement learning}
\ccsdesc[500]{Computer systems organization~Robotic control}
\ccsdesc[300]{Computing methodologies~Supervised learning by classification}
\ccsdesc[300]{Computing methodologies~Sequential decision making}
\ccsdesc[300]{Computing methodologies~Multi-task learning}

%%
%% Keywords. The author(s) should pick words that accurately describe
%% the work being presented. Separate the keywords with commas.
\keywords{Reinforcement Learning, Domain Adaptation, Damage recovery, Gait Selection, LSTM}

%%
%% This command processes the author and affiliation and title
%% information and builds the first part of the formatted document.
\maketitle

\section{Introduction}
One of the motives of introducing was to provide a safe method of access and operation in environments that are hazardous and unreachable to humans. But very often, these environments destabilize or damage the robot partially, often impairing them, and thus leading to a mission failure or significant drop in performance. This is especially critical for robots deployed in manufacturing industries and warehouses \cite{manuf}, search and rescue missions \cite{search} and disaster response \cite{disaster}. Although this situation of partial damage is tackled in humans or animals by their learning of alternate ways to perform the action, this kind of learning in robots requires, what we call, intelligence. Hence, the objective while designing robotic devices is not just restricted to avoiding or tackling obstacles, it also includes the adaptation of the agent in presence of adversaries, both in the form of internal damages as well as external effects.

Deep Reinforcement learning (Deep RL) has been shown to be effective in modeling such navigation problems because of both its online and offline learning capabilities in high dimensional search spaces \cite{rte, unfroze, rarl, visnav}. In the context of adapting to damages, offline learning would mean training a robust policy before the robot is deployed while online capabilities suggest learning to adapt at the time of damage.  But the environments and agents in these situations are very complex in nature, as a result of which, retraining the RL policy every time a change occurs in either of them is highly impractical. This points to the necessity of having an efficient control architecture which can help the agent adapt in varying adversarial conditions.

To implement this, several approaches have tried to learn multiple policies at training time and then choosing from them at the time of damage. However, models which have made progress in this domain require reset of the agent to initial state \cite{ite}, or multiple hardware trials are to be performed to help the agent recover or adapt \cite{ite, bongard, koos}. Although this is intuitive, it is inefficient considering the overhead of choosing from a set of high performing gaits. To make a smart recovery decision, an alternative method can be for agent to understand the damage first and then use that damage awareness to act optimally.

We thus propose Damage Aware-Proximal Policy Optimization (DA-PPO), combining damage diagnosis with deep reinforcement learning. The control architecture first performs damage diagnosis on multiple damage cases using a Long Short Term Memory (LSTM) \cite{lstm}, based supervised learning network. It uses the difference between the gaits of a (simulated) healthy and a damaged robot as input and classifies the damage that has occurred, if any. The data of the diagnosed damage is combined along with the current observation vector to create an augmented observation space, which contains information of both state space observation as well as damage. This augmented observation is used to train our RL model, which is optimized using Proximal Policy Optimization (PPO) \cite{ppo}. The trained model shall be able to understand the damage that has occurred and choose its gait accordingly. Since, only a single policy is learnt, there is no overhead of storing and choosing between multiple policies, making our algorithm effective in real time.

% We intend to create an efficient control structure that can tackle single and multiple internal damages in robotic agents in real time. 
The major objectives of our work are:
\begin{enumerate}
    \item To create a deep reinforcement learning based control architecture for enabling locomotory agents to accomplish mission-critical tasks even in the presence of single or multiple internal damages.
    \item To optimize the control architecture so that the agent adapts its gait in a single hardware trial.
\end{enumerate}

\section{RELATED WORK}

\subsection{Automated Recovery in Robotics}

Preliminary work on automated recovery in robots were based on evolutionary algorithms and generally divided the process into two phases - damage estimation and recovery. This necessitated the need for a healthy robot's simulation to always be available to the physical robot, so as to estimate the damage. This estimation would then help create a neural controller, during the exploration or recovery phase, which can handle the damage. The neural controller is passed to the robot and thus used for adaptation. The algorithm introduced in \cite{bongard}, was one of the first to propose an automatic and continuous information flow between a physical robot and its simulation wherein the robot provides its current state information. The simulator updates its own state using this information and provides the robot with neural controllers to handle its state or damage. The major advantage was that the creation of recovery method didn't have to be performed directly on the physical robot and thus the number of trials required to recover was drastically reduced.

Extending this work, Koos et al. \cite{koos}, also created a self diagnosis model. The main difference between the two works is the use of an undamaged self-model of the robot to find out behaviors rather than constantly updating it based on diagnosis. Although the intuition behind these are correct and applicable even today, the use of evolutionary algorithms make these methods inefficient.

\subsection{Map-based Algorithms for Adaptation}

Algorithms based on behavior performance maps \cite{ite, rte} rely on the assumption that knowledge of the cause of damage i.e., a proper diagnosis report is not necessary to recover from the damage. Rather than considering two separate phases for damage diagnosis and recovery algorithm generation, Cully et al. \cite{ite}, proposed a method inspired from animals, who perform trial and error to determine the least painful alternate gait in the presence of injury. The approach put forward in this work, Intelligent Trial and Error (ITE), relies on a behavior-performance map space. This map enables the robot to try multiple behaviors which are predicted to perform well. Based on the trials conducted and their results, the estimated  performance values are also updated in the map. The process converges when the best behavior possible has been estimated. Even when the damage is absent, the high performing behaviors are expected to be useful.

The implementation of this idea uses gaussian process \cite{rasmussen2005gaussian}, and  bayesian optimization procedure \cite{bayesian1,bayesian2}, to choose which gaits or behaviors to try at the time of damage by maximizing performance function from the  behavior-performance space. The selected gait is tested on the robot and its performance is recorded which then helps update the value of expected performance of that gait. This select-test-update loop continues till the right behavior is obtained.

Inspired by ITE, Chatzilygeroudis et al. \cite{rte}, proposed a more optimized version of the algorithm. Reset free trial and error (RTE) focuses on the fact that some of the high performing policies which work on an intact robot should also work on a damaged robot, which is true mainly in complex robotic systems like humanoids or multi-legged robots. Similar to ITE, RTE pre-computes and generates a behavior performance map using MAP-ELITES \cite{mapelites2015}. It learns the robot's model, especially when it is damaged and uses Monte Carlo Tree Search \cite{mcts2008} to compute the next best action for the current state of robot. Also, the method uses a probabilistic model to incorporate uncertainty of predictions and uses this data to correct the outcome of each action on the damaged robot \cite{alphago, texplore}. This culmination of algorithms makes sure that there is no reset required when a damage occurs.

A significant drawback in the previous two methods is the huge complexity overhead due to the use of gaussian process and also the inability to work on dynamic unknown terrains.

\subsection{Handling Environmental Adversaries}

Adversarial forces on robots are not limited to physical damages. There could also be environmental factors which hinder normal robotic locomotion. Several methods have been proposed to deal with these kind of damages. Robust Adversarial Reinforcement Learning (RARL) \cite{pinto2017icml}, concentrates on ensuring stability of an agent in the presence of an adversary, which is trying to destabilize it. It is based on the assumption that environmental changes, such as change in coefficient of friction of floor, between training and testing can also be modelled as an adversary acting on some part of the agent's body.

The algorithm is basically reduced to a min-max game where the adversary tries to minimize the reward of the concerned Markov Decision Process (MDP) and the protagonist tries to maximize it. The method of achieving this, as proposed, is to alternate between training of policies for both adversary and protagonist for a fixed number of iterations until convergence.

Another approach, introduced in \cite{kume}, is based on enabling adaptation to both environmental adversaries as well as physical or internal damage of robot. The major difference between their work and previous works like ITE and RTE is the existence of a multi-policy mapping for a single behavior in place of a single policy. Map-based Multi-Policy Reinforcement Learning (MMPRL), proposed in this work, trains many different policies by collaborating a behavior-performance map and the concepts of deep reinforcement learning. It aims to search and store these multiple policies while maximizing expected reward. MMPRL saves all possible policies with different behavioral features, making it extremely fast and adaptable.
% \newline
\vspace{-6pt}
\subsection{Domain Randomization}
Some recent works have also experimented with randomization in simulation environments through domain and dynamics randomization \cite{domain, dynam}, so as to bridge the gap between simulation and real world. The idea is to create numerous variations in the simulation environment so that real world appears as just another sample from a rich distribution of training samples. In \cite{domain}, the authors have experimented on object localization for the purpose of grasping in cluttered environment. They have shown impressive results, randomizing in the visual domain to transfer learning from simulation to real world without requiring real world training images.  On the other hand, in \cite{dynam}, the authors have randomized the dynamics of the environment such as mass, damping factor, friction coefficient and have shown that the policy learned in such a dynamic environment is quite robust to calibration errors in the real world. 

While most map-based methods are able to adapt over a wide range of damages, their computational overhead in creating the behaviour-performance map is a significant drawback. In ITE and RTE, the complexity is further increased by the gaussian process computations. Moreover, all these approaches require multiple hardware trials for adapting to a damage. We try to incorporate domain randomization approach in the context of damages so that damages in the real world are just another variation of training samples. Moreover, we further improve this approach by presenting a single hardware trial control loop for diagnosing the damage.
\vspace{-3pt}
\section{APPROACH}
\subsection{Overview}
We consider the following scenario: A robot has been damaged while in a remote and hazardous environment. We require the robot to reach the destination by adapting its gait so as to overcome the damage. Rather than making the agent dependent on a pre-computed set of high performing gaits, it should be able to identify and adapt to its damage autonomously.

Thus we propose a self-diagnose network which can predict the type of damage that has occurred in the structure of the robot. With this damage awareness, we use an augmented observation space for learning a well-performing policy through a modified version of Proximal Policy Optimization (PPO) which we call Damage Aware-Proximal Policy Optimization (DA-PPO). In our work, we assume that internal damages, unlike environmental adversaries, do not keep changing constantly. Thus, we only need to perform the self-diagnosis step for determining damage class whenever the reward drastically drops below a certain threshold, indicating that damage has occurred.

\subsection{Self-Diagnose Network}
\begin{figure}[h]
\centering
\includegraphics[width=0.40\textwidth]{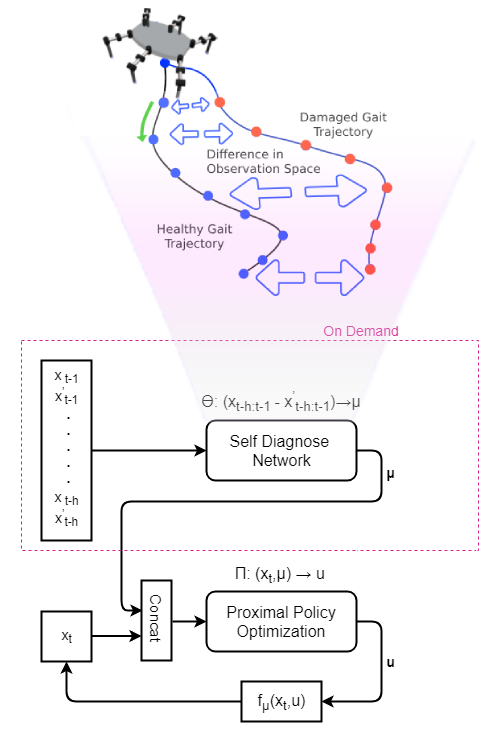}
\caption{Control Architecture}
\label{fig:ppoim}
\end{figure}

In the min-max based game approach put forward in RARL \cite{pinto2017icml}, the technique fails to generalize over changing damages from adversary at every time step. This is actually a tough problem since the policy has no feedback mechanism to judge the performance of action taken in the last time step.
% Thus we propose an LSTM based supervised learning neural network setup \cite{lstm1997} (see Figure \ref{fig:lstm}), which tries to guess the damage class of each motor joint.The idea here is that, the on board computer of the robot can run a healthy bot simulation and compare it with the actual steps taken. Based on the difference between the two, it can diagnose its class of damage.%
Thus we propose a self-diagnose network, an LSTM \cite{lstm} based predictive model, which tries to classify the type of damage that has occurred in the robot using continuous feedback from its gait. In \cite{bongard}, the authors have used the difference between the behaviours of simulated robot and the physical robot in terms of forward displacement to classify damages. We extend this idea by measuring the difference in sensor values between the two for a fixed number of time steps. This results in a time series and our problem is reduced to classifying damage from this data. More specifically, the on-board computer of the robot can run a simulation of a healthy robot and compare its gait with the actual steps taken. Based on the difference between the two, the network can diagnose the class of damage (see Fig. \ref{fig:ppoim}).

\begin{algorithm}

\SetAlgoLined
\KwResult{An array with collected samples}
Initialize: \\
Load an expert policy trained on healthy robot \\
Run parallel threads\\
 \For{$i\gets0$ \KwTo $n\_rollouts$ }{
    Set a random seed\\
    Initialize environments $env_h$, $env_d$ for healthy and damaged robots with the  same seed value\\
    \For{$damage\_class\gets0$ \KwTo $n\_damage\_classes$ }{
        $env_d$.applyDamage(damage\_class) \\
        \For{$n\gets0$ \KwTo $n\_timesteps$ }{
            get action from predefined policy\\
            $a_h$ = policy\_fn($obs_h$)\\
            $a_d$ = policy\_fn($obs_d$)\\
            do simulation step in both environments\\
            $obs_d$, $rew_d$ = $env_d$.step($a_d$)\\
            $obs_h$, $rew_h$ = $env_h$.step($a_h$)\\
            }
        collect ($o_h$-$o_d$)
        }
    }
    Concatenate collected samples
    \vspace{1ex}
 \caption{Sample collection}
 \label{alg:algo1}

\end{algorithm}

Since this time series is multivariate and high dimensional, we use LSTM hidden units which are powerful and increasingly popular models for learning from sequencial data \cite{surveylstm}. Algorithm \ref{alg:algo1} describes in detail the sample collection process. The healthy and damaged robot environments are represented by $env_h$ and $env_d$ respectively. Both the environments are run from the same initial state and the difference between their observation spaces is collected continuously for a fixed number of time steps $\mathit{T}$ (represented here as $n\_timesteps$). For any environment, this results in a matrix of size $\mathit{O} * \mathit{T}$ (where $\mathit{O}$ is the observation space size for that environment) and this represents a single data point. These data points act as training data, for which labels are the corresponding damage classes upon which the simulation was run. The whole process is repeated $n\_rollouts$ number of times to get multiple data points. Note that $policy\_fn$ represents an expert policy which has been pretrained on a healthy robot. 

The network is trained using data obtained through the sample collection step explained in Algorithm \ref{alg:algo1}. This step is also parallelized and thus doesn't act as a bottleneck for the entire algorithm. The self-diagnose network, represented by $\Theta$, can be accessed on demand to determine damage class $\mu$ within a single trial as shown in Fig. \ref{fig:ppoim}.

% \begin{algorithm}
% \SetAlgoLined
% \KwResult{Damage status for each joint of the robot}
% \begin{enumerate}
%   \item Load a batch from collected samples from algorithm 1
%   \item Create a neural network with one LSTM layer, three Dense layers and a softmax output with $n\_damage\_classes$
%   \item Train the neural network with categorical crossentropy loss
%   \item Repeat steps 1 to 3 until convergence
% \end{enumerate}
%  \caption{Self-diagnose Network}
% \label{alg:algo2}
% \end{algorithm}

\subsection{Encoding of Damage Indicators}

The self-diagnose network predicts the damage class of the robot which can act as an additional state information about the environment. We thus concatenate it with the observation space of the original robot to form, what we call, an augmented observation space.

This poses a necessity to encode the output of the classifier so that the policy efficiently learns various gaits in accordance with the damage. If a random encoding scheme is used for creating the augmented observation space, it results in the algorithm perceiving the encoding as noise, and completely ignoring it during policy learning. We have thus used partial one hot encoding and it is observed to work well in practice as the damage information is not lost during training.

In our experiments, we have limited the number of damages that can occur simultaneously to two and have taken the assumption that only one damage can occur on a limb at a time. The number of damage classes can thus be calculated as the sum of no damage case, single damage cases and multiple damage cases occurring at various limbs. This is given by:
\begin{equation}
\label{eqbin}
D =  k^0\binom{n}{0} +  k^1\binom{n}{1} +  k^2\binom{n}{2},
\end{equation}\\
where $n$ represents the number of limbs in the agent and $k$ represents the number of different damage types considered.

The encoded vector is of length $2n$ where the damage of $i^{th}$ limb is represented by the values at indices $2i$ and $2i+1$ in the encoded vector. Thus, we have a tuple of size 2 associated with each limb where $[0,0]$ represents no damage, $[1,0]$ represents damage type 1 and $[0,1]$ represents damage type 2 at the limb. Note that the tuple $[1,1]$ can be used if we remove the assumption that two types of damages can't occur together at a single limb. Furthermore, the tuple size can be increased to model more types of damages.

\subsection{Proximal Policy Optimization}
Since our task is that of continuous action control, we formulate it as a reinforcement learning problem, starting from initial state $s_0$, choosing a series of action $a$ and obtaining state $s_i$ and reward $r_i$ at the $i^{th}$ timestep while maximizing the expected sum of rewards by changing the parameter $\theta$ of the parameterized stochastic policy $\pi_\theta$. But the use of large scale optimization is less widespread in continuous action spaces. An attractive option for such problems is to use policy gradient algorithms \cite{poli}. Proximal Policy Optimization is a simplified version of Trust Region Policy Optimization (TRPO) \cite{trpo}. It improves upon the stability of policy gradient methods by allowing multiple updates on minibatch of on-policy data. This is implemented by limiting the KL divergence between updated policy and the policy from which the data was sampled. TRPO uses a hard optimization constraint for achieving the same but is computationally expensive to compute. PPO approximates TRPO by using a soft constraint. The original paper \cite{ppo} proposes two methods for implementing this soft constraint: an adaptive KL loss penalty and using a clipped surrogate loss function.

PPO represents the ratio between new policy and old policy as:
\begin{equation} \label{eq1}
    r_t(\theta) = \frac{\pi_\theta(a_t | s_t)}{{\pi_\theta}_\textsubscript{old}(a_t | s_t)} .
\end{equation}

The objective functions can be \cite{ppo}:
\begin{equation}
    L\textsuperscript{CLIP}(\theta) = \mathbb{\hat{E}}_t[\text{min}(r_t(\theta)\hat{A}_t, \text{clip}(r_t(\theta), 1 - \epsilon, 1 + \epsilon)\hat{A}_t)] ,
    \label{eqclip}
\end{equation}
\begin{equation}
    L\textsuperscript{KL}(\theta) =
    r_t(\theta)\hat{A}_t - \beta{KL}[{\pi_{\theta_{old}}}, \pi_{\theta}] 
    \label{eqkl}
\end{equation}
Both these objective functions stabilize training by constraining the policy changes at each step, thus approximating the gradient to a local value, so that large steps are not taken between iterations.
Additionally, we use Generalized Advantage Estimation (GAE) \cite{schulman2015high} for computing the advantage function $\hat{A}$. In our implementation of PPO, we have used a combination of both clipping loss and adaptive KL penalty for locomotion tasks. The hyperparameters for the same are mentioned in Section \ref{netset}.
% \begin{algorithm}[H]
% \SetAlgoLined
% \begin{algorithmic}[1]
%     \FOR{iteration=1,2... }
%       \FOR{actor=1,2,...,N}
       
%   \item[] Run policy $\pi$ in environment for T timesteps
%   \item[] Compute advantage estimates
%   \ENDFOR
%   \STATE Optimize surrogate $\mathit{L}$
%   wrt $\theta$, with $\mathit{K}$ epochs and\\ minibatch $\mathit{M}$ $\leq$ $\mathit{L*T}$
%  \STATE $\theta$\textsubscript{old} $\leftarrow$ $\theta$
%   \ENDFOR
%   \end{algorithmic}

%  \caption{PPO, Actor-Critic Style}
%  \label{alg:algo3}
% \end{algorithm}

% In actor-critic PPO implementation, \cite{ppo}, $\mathit{N*T}$ samples are collected initially, where $\mathit{T}$ is the number of rollout steps each actor takes between each update and $\mathit{N}$ is the number of such actors. From these rollouts, generalized advantage estimates \cite{advantage} are calculated.

% Mini-batches of size $\mathit{m}$ are sampled from the $\mathit{N*T}$ sized batch. Initially the ratio at equation \ref{eq1} is set to 1. After the update from the first mini-batch of first epoch occurs, clipping is performed. Updates are performed on each individual mini-batches and are shuffled to generate a new combination in the next epoch. This is repeated for $\mathit{K}$ epochs before the old parameters are updated to reflect the current values.
\subsection{Damage Aware Proximal Policy Optimization}
With the self-diagnose network in place, we can now use the policy learning algorithm on augmented observation space which encapsulates both environment state (through observation vector) and damage awareness (through damage encoding vector). We use the PPO algorithm for policy learning from the augmented observation space where $x_t$ is the observation at timestep $t$, $u$ is the action taken according to policy $\Pi$ and $f_\mu$ is the environment in which damage $\mu$ has occurred (see Fig. \ref{fig:ppoim}). Note that we only run self-diagnose network when reward during a run falls below a certain threshold. At other times, the damage is considered to be the same as diagnosed in the last run.

% To define the indicators for the ant environment, for example, we have considered 2 damage classes acting on a total of 4 joints, i.e., all four hip joints of the Ant robot (see Figure \ref{fig:single_damage}. The damages can also be multiple with maximum of 2 damages occurring simultaneously (see Figure \ref{fig:multiple_damages}). The case of a healthy robot, i.e., no damage, is also considered. One-hot encoding has been to encode an agent's damage, represented by a vector of size 8, where each sequence of 2 elements denote the damage class in one joint. This results in a total of 33 possible damage class combinations.

\section{Experimental Setup}

\subsection{Simulation Setup}
To evaluate our approach, we have conducted experiments on two environments, \textit{Ant}, a quadrupedal locomotory robot and \textit{Hexapod}, a six-legged locomotory robot. We have used OpenAI gym toolkit \cite{gym}, for performing simulations in combination with MuJoCo physics engine \cite{mujoco}. The Ant is an already implemented environment in OpenAI Gym while the Hexapod is implemented using the configuration and model described in ITE \cite{ite}.

The two environments used in our experiments are discussed below:\\
% \begin{figure}[h]
% \centering
% \begin{center}
% \begin{subfigure}{0.24\textwidth}
% \includegraphics[width=0.9\linewidth,height=3.25cm]{images/antpoorbig.png}
% \caption{Missing toe 1}
% \label{fig:sideways}
% \end{subfigure}%
% \begin{subfigure}{0.24\textwidth}
% \includegraphics[width=0.9\linewidth, height=3.5cm]{images/hexapoorbig.jpg}
% \caption{Missing toe 2}
% \label{fig:frontways}
% \end{subfigure}
% \end{center}
% \caption{Single damage classes showing missing lower limb (\ref{fig:sideways}, \ref{fig:frontways}) and jammed upper limb joint (\ref{fig:joint1}, \ref{fig:joint2})}
% \label{fig:single_damage}
% \end{figure}

\textbf{Ant (Quadrupedal bot)}: Ant is a simple quadrupedal robot with 12 degrees of freedom (DoF) and 8 torque actuated joints. The joint has maximum flex and extension of 30 degrees from their original setting and also has a force and torque sensor. The observation includes features containing joint angles, angular velocity, the position of all structural elements with respect to the center of mass and force and torque sensor outputs of each joint forming a 111-dimension vector. The target action values are the motor torque values which are limited in the range -1.0 to 1.0. We limit an episode to at most 1000 timesteps and the episode will end whenever it crosses this limit or robot falls down on its legs or jumps above a certain height.
The reward function is defined as follows:
\begin{equation}
    R_t = \Delta x_t + s_t - w_0C_t -(w_1||\phi_t||_2)^2,
\end{equation}
where \(\Delta x_t\) is the covered distance of the robot in the current time step since the previous time step, \(s_t\) is the  survival reward, which is 1 on survival and 0 if the episode is terminated by the aforementioned conditions. The variable \(C_t\) is the number of legs making contact with the ground, \(\phi_t \in R^{8}\)  are the target joint angles (the actions),
and $w_n$ is the weight of each component with $w_0$ = 0.5, $w_1$ = 0.5.\\
% When forming an augmented observation space, an additional 8-dimension damage encoding vector is concatenated to form a 119-dimension observation vector.

\textbf{Hexapod}: There are three actuators on each leg of the Hexapod. %%(see Fig. \ref{fig:single_damage})%%.
In the neutral position, the height of the robot is 0.2 meters. In addition to this, the actions are taken to be the joint angle positions of all 18 joints, which ranges from -0.785 to 0.785 radians.
As the observation space of the agent, a 53-dimension vector is given as input which consists of the position and velocity of all the joints as well as he center of mass. Along with this, the observation space contains boolean values from touch sensors which indicate whether a leg is making contact with the ground or not. Again, we limit an episode to be at most 1000 timesteps and the episode will end whenever the robot falls down on its legs or jumps above a certain height or crosses the time limit.

The reward function R is defined as follows:
\begin{equation}
    R_t = \Delta x_t + s_t - w_0C_t - (w_1||\tau_t||_2)^2 - (w_2||\phi_t||_2)^2,
\end{equation}
where \(\Delta x_t\) is the covered distance of the robot in the current time step since the previous time step, \(s_t\) is the  survival reward, which is 0.1 on survival and 0 if the episode is terminated by the aforementioned conditions. The variable \(C_t\) represents the number of legs making contact with the ground, \(\tau_t \in R^{18}\) is the vector of squared sum of external forces and torques on each joint, \(\phi_t \in R^{18}\)  are the target joint angles (the actions), and $w_n$ is the weight of each component with $w_0$ = 0.03, $w_1$ = 0.0005, and $w_2$ = 0.05.
% Again, for forming an augmented observation space, we concatenate a 12-dimension damage encoding vector to form a 65-dimension observation vector.

% \begin{figure}[h]

% \begin{subfigure}{0.25\textwidth}
% \includegraphics[width=0.9\linewidth, height=4cm]{images/ant.png}
% \caption{Ant}
% \label{fig:ant}
% \end{subfigure}%
% \begin{subfigure}{0.25\textwidth}
% \includegraphics[width=0.9\linewidth, height=4cm]{images/hexapod.png}
% \caption{Hexapod}
% \label{fig:hexapod}
% \end{subfigure}
% \captionsetup{justification=centering}
% \caption{Agent Environments\\(Source (a) and (b):  OpenAI Gym \cite{brockman2016gym}, Source(c): Robots that can adapt like animals \cite{antoine2015nature})}
% \label{fig:agents}
% \end{figure}

\subsection{Damage Simulation}
Since both the environments considered in our experiments are simulated in OpenAI gym, the damages are implemented by changing the \textit{xml} files of the 3D models. This can be done on the fly without affecting parallely running experiments. In our work, we have simulated broadly two kinds of internal damages which are
\begin{enumerate}
    \item Jamming of joint such that it can't move irrespective of the amount of torsional force applied by the motor at that joint.
    \item Missing toe, i.e., lower limb of the robot breaks off.
\end{enumerate}

\begin{figure}
\centering
\begin{subfigure}{0.25\textwidth}
\includegraphics[width=0.9\linewidth,height=3.3cm]{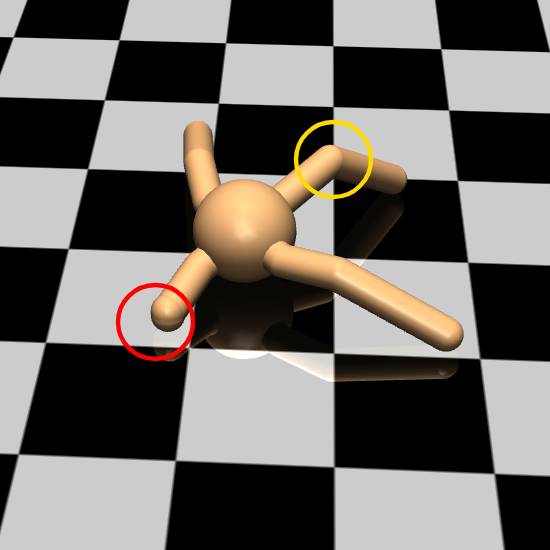}
\caption{Ant damage scenario 1}
\label{fig:sideways}
\end{subfigure}%
\begin{subfigure}{0.25\textwidth}
\includegraphics[width=0.9\linewidth, height=3.3cm]{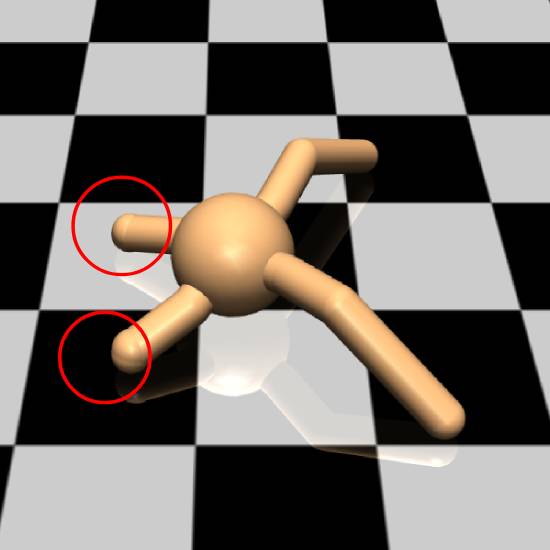}
\caption{Ant damage scenario 2}
\label{fig:frontways}
\end{subfigure}
\begin{subfigure}{0.25\textwidth}
\includegraphics[width=0.9\linewidth,height=3.3cm]{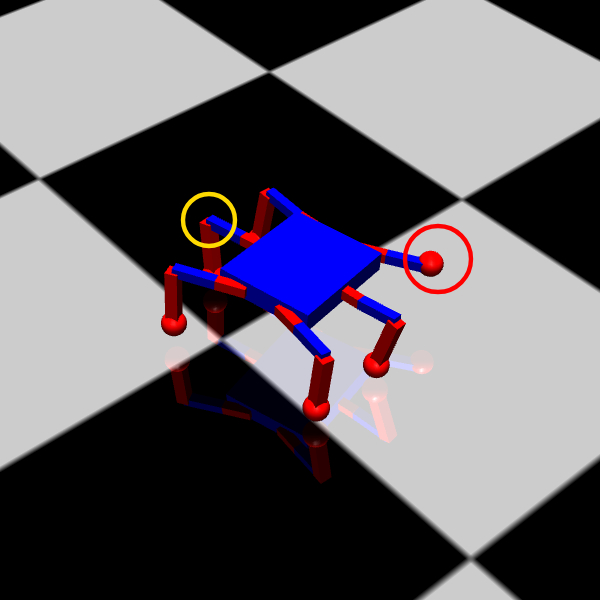}
\caption{Hexapod damage scenario 1}
\label{fig:joint1}
\end{subfigure}%
\begin{subfigure}{0.25\textwidth}
\includegraphics[width=0.9\linewidth, height=3.3cm]{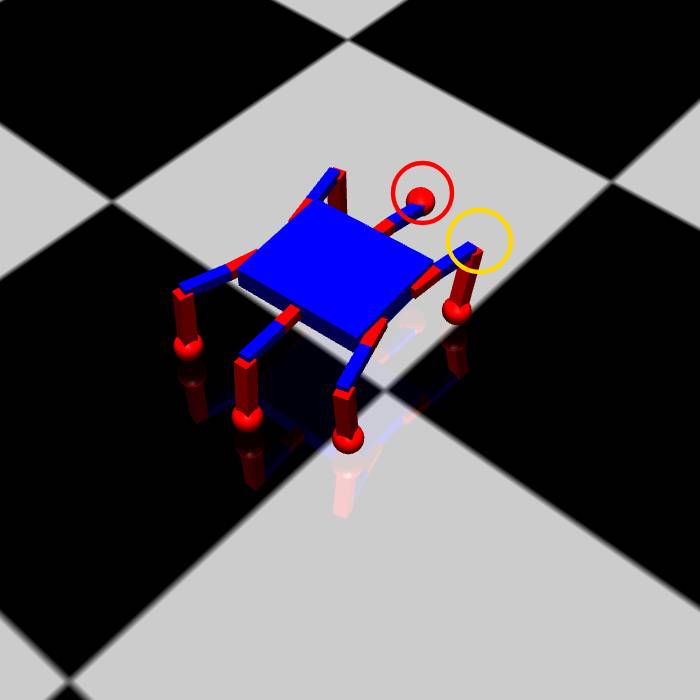}
\caption{Hexapod damage scenario 2}
\label{fig:joint2}
\end{subfigure}
\caption{Some of the damage scenarios in Ant and Hexapod. Yellow and red circles represent jammed joint and missing limb damage types respectively.}
\label{fig:single_damage}
\end{figure}

In MuJoCo environments, these damages are implemented as follows:

    Ant Environment
\begin{itemize}

\item Jamming of joint is  modelled by restricting the angle range of concerned joint  to -0.1 to 0.1 degrees from the default value of -30 to 30 degrees.

\item Missing toe is modelled by shrinking the lower limb size to 0.01 from the original value of 0.8.
\end{itemize}

    % \item Hexapod Environment
    Hexapod Environment
\begin{itemize}
\item The original angle range of hexapod is -45 to 45 degrees. This is restricted to -0.1 to 0.1 when jamming of joint is modelled.

\item Missing of any of the 6 toes in hexapod is modelled by reducing the the lower limb size to 0.01 instead of 0.07 in healthy robot.
\item There are touch sensors on each lower limb of the hexapod. Thus  whenever a lower limb breaks off we consider that the touch sensor corresponding to it stops giving any signal and its output is considered to be 0. 
\end{itemize}

\begin{figure*}[h!]  % spans both columns
\begin{subfigure}{0.49\textwidth}
\includegraphics[width=\linewidth]{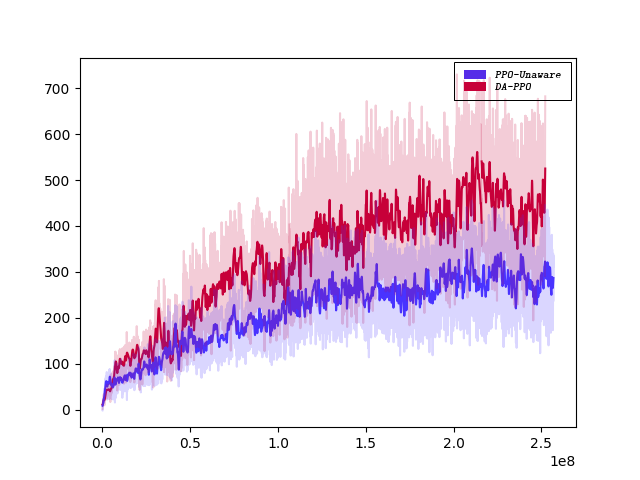}
\caption{Ant Environment}
\label{fig:anttrain}
\end{subfigure}
\hfill % maximize the horizontal distance between the graphs
\begin{subfigure}{0.49\textwidth}
\includegraphics[width=\linewidth]{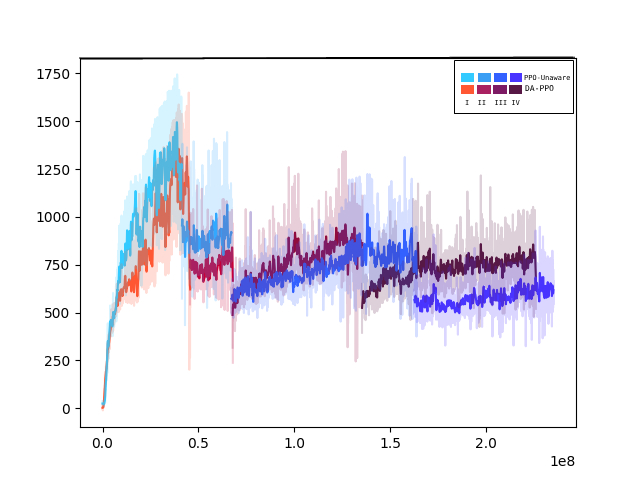}
\caption{Hexapod Environment}
\label{fig:hextrain}
\end{subfigure}
\caption{Training curve comparision between DA-PPO and PPO Unaware in both Ant and Hexapod Environments}
\label{fig:curve}
\end{figure*}
\subsection{Hyperparameter Details} \label{netset}
For the self-diagnose network, we take as input a  matrix of size $batch\_size *\mathit{O} * \mathit{T}$  and this is followed by an embedding layer with embedding size 512  and an LSTM layer with 32 hidden units. After this, we stack three dense layers of size 256, 128, 64 along with dropouts, so as to reduce overfitting. The output layer uses \textit{softmax} as activation so that it outputs class probabilities. The loss function and optimizer used are \textit{categorical crossentropy} and \textit{adam} respectively. For Ant and Hexapod environments, the possible classes range from 0 to 32 and 0 to 72 respectively as calculated from Equation \ref{eqbin}.

As for the policy learning using PPO, we use the implementation from \cite{TFAgents}. For both value function and policy function, we use the same network configuration having hidden layer sizes as 100, 200, 100. Adam optimizer was used for both the neural networks. The GAE gamma value is taken as 0.995 and lambda as 0.98. The clipping range is kept at 0.2 and adaptive KL target is initialized with 0.01. Adam learning rate and KL target value are adjusted dynamically during the training. Moreover, we trained the value function on the combination of current batch and previous batch to stabilize training.
\section{Results and Discussion}

We evaluate the performance of our approach within the two elements involved :
(1) Self-Diagnose network for predicting class of damage (2) DA-PPO, which learns to adopt a policy given that a particular damage has occurred.

\subsection{Self-Diagnose Network}
For the comparison of performance, we consider different number of rollouts (amount of data to train on), length of history to look back into (timesteps) and what to give as observation data, i.e., our proposed approach of using difference of observations between healthy and damaged run or the observations of only damaged run. Table \ref{table:1} summarizes the validation accuracy across these parameters. We can observe that classifying using fewer timesteps results in faster diagnosis but at the expense of accuracy. Moreover, classification using the difference between observation vectors as input outperforms the use of observations from damaged run only in all the cases. However, if there is a constraint on computation power of the on-board computer of the robot, the latter approach can be preferred over the former one.

\begin{table}[h!]
\caption{Classification accuracy in predicting damage class in Ant and Hexapod environment with varying number of timesteps and rollouts. Method A represents using observations of damaged run only as time series and method B represents using difference of observations between healthy robot and damaged robot as time series.}
\resizebox{\columnwidth}{!}{
\begin{tabular}{|c|c|c|c|c|}
% \hline
\multicolumn{5}{c}{Classification Accuracy in Ant Environment}                                                         \\ \hline
\multirow{2}{*}{Timesteps} & \multirow{2}{*}{Method} & \multicolumn{3}{c|}{Number of Rollouts}                           \\ \cline{3-5} 
                           &                         & 1000                 & 2000                & 7000                 \\ \hline
\multirow{2}{*}{10}        & A                       & 78.2$\pm$1.11           & 81.4$\pm$0.6           & 82.4$\pm$0.87           \\ \cline{2-5} 
                           & B                       & \textbf{81.24$\pm$2.88} & \textbf{85.2$\pm$1.2}  & \textbf{84.33$\pm$0.72} \\ \hline
\multirow{2}{*}{30}        & A                       & 82.17$\pm$1.7           & 87.1$\pm$1.8           & 88.17$\pm$1.3           \\ \cline{2-5} 
                           & B                       & \textbf{83.62$\pm$2.03} & \textbf{90.8$\pm$0.9}  & \textbf{91.5$\pm$1.067} \\ \hline
\multirow{2}{*}{50}        & A                       & 83.11$\pm$0.8           & 90.17$\pm$1.2          & 92.83$\pm$1.8           \\ \cline{2-5} 
                           & B                       & \textbf{84.29$\pm$1.21} & \textbf{92.6$\pm$1.83} & \textbf{96.8$\pm$1.48}  \\ \hline
\end{tabular} }
% \vspace{-13pt}
\label{table:1}
% \end{table}
% \quad
% \begin{table}[h]
% \caption{Classification accuracy in predicting damage class in Hexapod environment with varying number of timesteps and rollouts. Method A represents using plain observations as time series and method B represents using difference of observations between healthy robot and damaged robot as time series.}
% \newline

% \newline
\resizebox{\columnwidth}{!}{
\begin{tabular}{|c|c|c|c|c|}
% \hline
\multicolumn{5}{c}{Classification Accuracy in Hexapod Environment}                                                    \\ \hline
\multirow{2}{*}{Timesteps} & \multirow{2}{*}{Method} & \multicolumn{3}{c|}{Number of Rollouts}                          \\ \cline{3-5} 
                           &                         & 1000                & 2000                 & 7000                \\ \hline
\multirow{2}{*}{10}        & A                       & 22.2$\pm$0.6           & 33.1$\pm$1.23           & 44.6$\pm$0.9           \\ \cline{2-5} 
                           & B                       & \textbf{32.6$\pm$0.8}  & \textbf{38.5$\pm$1.1}   & \textbf{47.8$\pm$1.13} \\ \hline
\multirow{2}{*}{30}        & A                       & 60.5$\pm$1.9           & 62.9$\pm$1.8            & 79.67$\pm$1.02         \\ \cline{2-5} 
                           & B                       & \textbf{65.45$\pm$1.2} & \textbf{69.17$\pm$1.11} & \textbf{82.6$\pm$1.28} \\ \hline
\multirow{2}{*}{50}        & A                       & 65.23$\pm$1.3          & 69.7$\pm$1.1            & 82.2$\pm$1.8           \\ \cline{2-5} 
                           & B                       & \textbf{68.83$\pm$1.8} & \textbf{72.17$\pm$1.29} & \textbf{87.6$\pm$0.86} \\ \hline
\end{tabular} }
\vspace{-10mm}
\label{table:2}
\vspace{6mm}
\end{table}
% \vspace{-15mm}
% Instead of feeding raw observations, actions and reward, the sample collection step gives training data to the diagnose network where the input is the difference between state space of damaged gait and undamaged gait simulations, and output is damage class label. Experiments have been done for different number of rollouts, varying length of timestep history for each prediction and different damage combinations.

% From the results of single limb damage, it is clearly visible that the model is unable to diagnose the damage when considering only a few timesteps of history, but has a huge increase in accuracy as this lookback parameter is increased. Accuracy is also seen to be directly proportional to the number of rollouts. It has been observed that the network is able to predict the class of damage with xxx accuracy for ant environment when run for 10000 rollouts and xxx damages (see table \ref{table:x}). This suggests that this approach can reliably diagnose a variety of damages. Also, it has been noted that the network is able to generalize over unseen damages upto an extend by mapping them as a class of damage with similar gait difference. This is suitable for our objective of adaptation as can be seen later.

% The results of these experiments are summarized in tables \ref{table:4}, \ref{table:5}, \ref{table:6}.
% \usepackage{multirow}
% \vspace{-2em}
\begin{figure*}
\includegraphics[width=\linewidth]{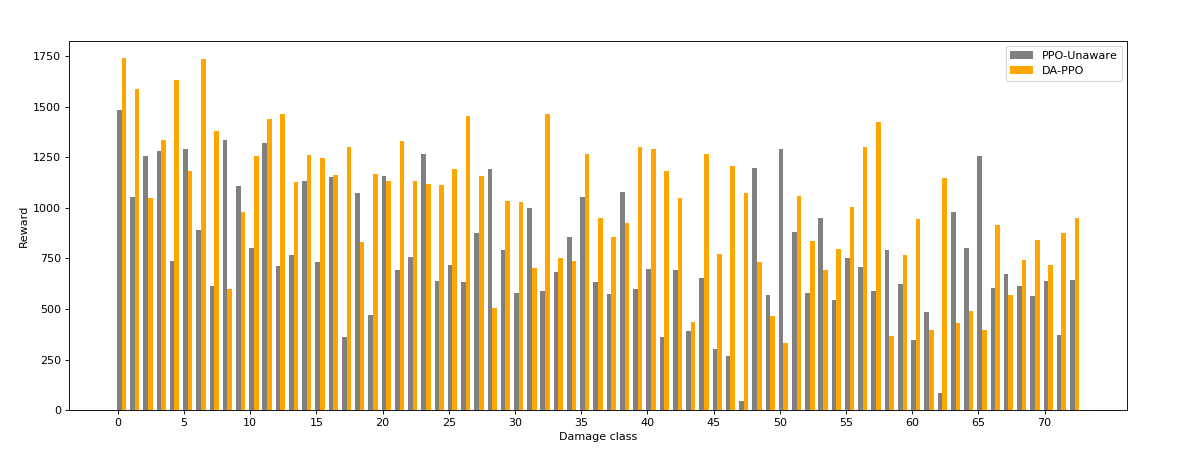}
\caption{Forward reward comparison between DA-PPO and PPO-Unaware across different damage classes in Hexapod} % Overall figure caption
\label{fig:comparehex}
\end{figure*}
\subsection{Damage Aware-Proximal Policy Optimization}
We start by creating a baseline model for comparison of performance. We define a model using PPO policy which is trained on experiments having damaged robot but without augmented observation space (i.e., without explicit knowledge of damage class), and call it PPO-Unaware. This is analogous to having a policy implementing domain randomization in damage space but without having a feedback loop. Our proposed model, which uses Damage Aware PPO policy, is called DA-PPO. The performance metric used is the forward reward of the agent, averaged across all the damage classes. Fig. \ref{fig:curve} shows the training curve comparison between PPO-Unaware and DA-PPO in Ant and Hexapod environments (see Fig.  \ref{fig:anttrain}, \ref{fig:hextrain} ). DA-PPO shows a 60.7\% improvement in average forward reward in Ant environment while in Hexapod environment, there is a 31.5\% reward gain over PPO-Unaware.

% \begin{figure}[h]
% \centering
% \begin{subfigure}{0.25\textwidth}
% \includegraphics[width=\linewidth,height=4cm]{images/anttrainingcurveedit.png}
% \caption{Training curve of Ant}
% \label{fig:anttrain}
% \end{subfigure}%
% \begin{subfigure}{0.25\textwidth}
% \includegraphics[width=\linewidth, height=4cm]{hexatrainingnice.jpg}
% \caption{Training curve of Hexapod}
% \label{fig:hextrain}
% \end{subfigure}
% \caption{Average reward comparison between DA-PPO and PPO-Unaware in Ant and Hexapod. In Hexapod, each piece-wise curve represents a stage (I, II,  II or IV)Iin the curriculum learning process. I has 100\% healthy cases, II has 60\% healthy and 40\% single damage cases, III has 70\% healthy and single damage and 30\% multiple damage cases and IV has all damages equally likely.  }
% \label{fig:curve}
% \end{figure}

\begin{figure}
\includegraphics[width=\linewidth]{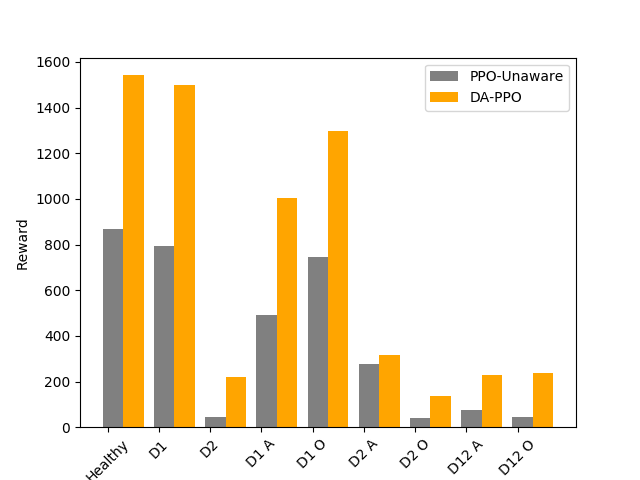}
\caption{Forward reward Comparison between PPO-Unaware and DA-PPO across different grouped damage classes in Ant. D1 and D2 refers to single jammed joint and single missing toe damages. ${D_i}_j$ A and O represents that damage type $i$ and $j$ are present in adjacent (A) or opposite (O) limbs. }
\label{fig:compareant}
\end{figure}
% The average forward reward in DA-PPO is 450 in Ant and 750 in Hexapod while in PPO-Unaware it is 250 in Ant and 600 in Hexapod. This shows that our method consistently performs better than training on damage without feedback.

For the Hexapod environment, we also use the concept of curriculum learning \cite{curr}, by progressively training on cases which are more difficult. We implement this by increasing the percentage of damage classes in training examples and also progressively increasing the severity of damages (by including multiple damages).
In Fig. \ref{fig:hextrain}, each piece-wise curve represents a stage (I, II,  III or IV) in the curriculum learning process. I has 100\% healthy cases, II has 60\% healthy and 40\% single damage cases, III has 70\% healthy and single damage cases and 30\% multiple damage cases and IV has all damages equally likely. In this way, we were able to encourage a faster learning progress.

% Although this poses a necessity to encode the output of the classifier so that our policy learns various gaits in accordance with the damage, efficiently. Any random encoding would result in the algorithm perceiving it as noise, and completely avoiding it. We have used partial one hot encoding and have seen that this encoding scheme encompasses the information of damage well and is not lost in training. Two kinds of damage have been considered with at most 2 simultaneous damages, resulting in a total of 33 damage classes.

% The 8 valued vector representing the damage denotes damage class for the 4 limbs as tuples of 2 per limb. This encoding scheme has been proven to be foolproof to improper encoding, where the reward drops to as much as half if a wrong encoding is given. This shows that the algorithm does use the damage information to compute the right gait.

We also do a per class performance analysis of the two approaches discussed across various damage classes in both Ant and Hexapod (see Fig. \ref{fig:comparehex}, \ref{fig:compareant}). In the Ant environment, DA-PPO performs better in 82.84\% of damage classes when compared to PPO-Unaware. Comparing between various damage classes, DA-PPO is seen to adapt really well (in terms of reward improvement over PPO-Unaware) when damages occur on opposite limbs as compared to damages occurring on adjacent limbs. In the Hexapod environment, DA-PPO performs better in 72.6\% of damage classes when compared to PPO-Unaware (see Fig. \ref{fig:comparehex}). This shows that being damage aware results in significant improvement in performance in presence of adversaries.

\section{CONCLUSIONS}

We have proposed and implemented a two-part control architecture for robotic damage adaptation. This is particularly useful when robots are used in hazardous environments, where human intervention is nearly impossible.

Our approach enables the agent to autonomously identify and understand the damage that has occurred in its physical structure and adapt its gait accordingly. Since the ultimate goal is the creation of intelligent machines, understanding the damage is as important as adapting from it, which has often been overlooked in past works.

On comparison with map-based approaches, DA-PPO doesn't require any map generation phase and thus the initial training time is much less. This is also enhanced by the fact that our approach adapts to the damage in a single trial itself, without trying multiple well-performing gaits or without having to be reset to the initial state to perform the trial.

Our work can also be easily scaled to a larger number of damage classes. Since no differentiation is made between the cause of damage, adaptation is possible in case of both morphological and external damages. Also, in the case of unknown damages, the network is expected to predict a damage class which resembles the actual damage the most and try to choose a gait accordingly. This implies a very low rate of complete failure. We intend to study more on this in a future work.

Future work shall be focused on extending the algorithm to handle environmental adversaries, which is much desirable since real-world environments are not predictable.  We also intend to work on DA-PPO for complex and dynamic environments, using SLAM \cite{slam}. Finally, we plan to extend our method and prove its effectiveness by applying it on a physical robot. 
\bibliographystyle{ACM-Reference-Format}
\bibliography{bibi}

% %%
% %% If your work has an appendix, this is the place to put it.
% \appendix

% \section{Research Methods}

% \subsection{Part One}

% Lorem ipsum dolor sit amet, consectetur adipiscing elit. Morbi
% malesuada, quam in pulvinar varius, metus nunc fermentum urna, id
% sollicitudin purus odio sit amet enim. Aliquam ullamcorper eu ipsum
% vel mollis. Curabitur quis dictum nisl. Phasellus vel semper risus, et
% lacinia dolor. Integer ultricies commodo sem nec semper.

% \subsection{Part Two}

% Etiam commodo feugiat nisl pulvinar pellentesque. Etiam auctor sodales
% ligula, non varius nibh pulvinar semper. Suspendisse nec lectus non
% ipsum convallis congue hendrerit vitae sapien. Donec at laoreet
% eros. Vivamus non purus placerat, scelerisque diam eu, cursus
% ante. Etiam aliquam tortor auctor efficitur mattis.

% \section{Online Resources}

% Nam id fermentum dui. Suspendisse sagittis tortor a nulla mollis, in
% pulvinar ex pretium. Sed interdum orci quis metus euismod, et sagittis
% enim maximus. Vestibulum gravida massa ut felis suscipit
% congue. Quisque mattis elit a risus ultrices commodo venenatis eget
% dui. Etiam sagittis eleifend elementum.

% Nam interdum magna at lectus dignissim, ac dignissim lorem
% rhoncus. Maecenas eu arcu ac neque placerat aliquam. Nunc pulvinar
% massa et mattis lacinia.

\end{document}